\documentclass[conference]{IEEEtran}
\usepackage{soul}

\usepackage{colortbl}
\usepackage{xspace}
\usepackage{todonotes}
\PassOptionsToPackage{hyphens}{url}\usepackage{hyperref}
\usepackage{csquotes}
\usepackage{mdwlist}
\usepackage{paralist}
\usepackage{tabularx}
\usepackage{subcaption}
\usepackage{graphicx}
\usepackage{amsmath,amssymb}
\usepackage{algorithm}
\usepackage{algpseudocode}
\usepackage{listings}
\usepackage{courier}
\usepackage{multirow}
\usepackage{booktabs}
\usepackage{balance} 
\usepackage{rotating}

\usepackage{enumitem}

\def\addauthnote#1#2{%
	\expandafter\def\csname#1\endcsname##1{\todo[inline,color=#2]{#1: 
		##1}\xspace}
	\expandafter\def\csname#1m\endcsname##1{\todo[color=#2]{#1: ##1}\xspace} 
}

\addauthnote{wil}{green!25} \addauthnote{azer}{blue!25}
\addauthnote{renato}{yellow!25}

\graphicspath{ {./figures/} }

\newcommand\eg{\emph{e.g.,}\xspace}
\newcommand\ie{\emph{i.e.,}\xspace}

\newcommand\gym{\textsc{GymFC-v1}\xspace}
\newcommand\newgym{\textsc{GymFC-v1.5}\xspace}
\newcommand\fc{Neuroflight\xspace}
\newcommand\depsize{24.86 KB\xspace}

\newcommand\ChNoiseYDeltaDecrease{87.95\%\xspace}

\newcommand\ChYDecrease{90.56\%\xspace}

\newcommand\rOmegaDeltaError{45.07\%\xspace}
\newcommand\rOmegaDeltaPower{3.41\%\xspace}

\newcommand\rPpoHorizon{500\xspace}
\newcommand\rPpoStepsize{$1 \times 10^{-4} \times \rho$\xspace}
\newcommand\rPpoEpochs{5\xspace}
\newcommand\rPpoMinibatch{32\xspace}
\newcommand\rPpoDiscount{0.99\xspace}
\newcommand\rPpoGae{0.95\xspace}

\newcommand\rGraphSize{12KB\xspace}
\newcommand\rOptGraphDecrease{16\%\xspace}

\newcommand\rTimeNfAlgDisarmedWCET{204\xspace}
\newcommand\rTimeNfAlgDisarmedBCET{194\xspace}
\newcommand\rTimeNfAlgDisarmedP{4.9\xspace}

\newcommand\rTimeNfAlgArmedWCET{210\xspace}
\newcommand\rTimeNfAlgArmedBCET{195\xspace}
\newcommand\rTimeNfAlgArmedP{7.1\xspace}

\newcommand\rTimeNfLoopDisarmedWCET{244\xspace}
\newcommand\rTimeNfLoopDisarmedBCET{229\xspace}
\newcommand\rTimeNfLoopDisarmedP{6.1\xspace}

\newcommand\rTimeNfLoopArmedWCET{423\xspace}
\newcommand\rTimeNfLoopArmedBCET{263\xspace}
\newcommand\rTimeNfLoopArmedP{37.8\xspace}

\newcommand\rTimeBfAlgDisarmedWCET{14\xspace}
\newcommand\rTimeBfAlgDisarmedBCET{9\xspace}
\newcommand\rTimeBfAlgDisarmedP{35.7\xspace}

\newcommand\rTimeBfAlgArmedWCET{15 \xspace}
\newcommand\rTimeBfAlgArmedBCET{9 \xspace}
\newcommand\rTimeBfAlgArmedP{40.0 \xspace}

\newcommand\rTimeBfLoopDisarmedWCET{58\xspace}
\newcommand\rTimeBfLoopDisarmedBCET{45\xspace}
\newcommand\rTimeBfLoopDisarmedP{22.4\xspace}

\newcommand\rTimeBfLoopArmedWCET{238\xspace}
\newcommand\rTimeBfLoopArmedBCET{78\xspace}
\newcommand\rTimeBfLoopArmedP{67.2\xspace}

\newcommand\rMaxNNTaskFreq{2.4kHz\xspace}

\newcommand\rLoopDenom{2\xspace}

\newcommand\rMaxLoopHertz{2kHz\xspace}
\newcommand\rMaxGyroHertz{4kHz\xspace}

\newcommand\rFasterThanPX{8\xspace}
\newcommand\rFastThanPWM{40\xspace}
\newcommand\rSlowerThanBF{1.78\xspace}

\newcommand\powerVoltage{16.78\xspace}
\newcommand\ampsNFdisarmed{0.37\xspace}
\newcommand\ampsNFarmed{0.67\xspace}

\newcommand\powerNFdisarmed{6.21\xspace}
\newcommand\powerNFarmed{11.24\xspace}

\newcommand\ampsBFdisarmed{0.37\xspace}
\newcommand\ampsBFarmed{0.6\xspace}

\newcommand\powerBFdisarmed{6.21\xspace}
\newcommand\powerBFarmed{10.07\xspace}

\newcommand\tunableWeights{1344\xspace}

\newcommand\aircraft{NF1\xspace}

\newcommand\nn{NN\xspace}
\newcommand\nns{NNs\xspace}

\newcommand\videourl{\url{https://wfk.io/neuroflight}\xspace}

\newcommand\sourcecode{\url{https://github.com/wil3/neuroflight}\xspace}

\soulregister{\CHone}{7}
\soulregister{\CHtwo}{7}
\soulregister{\CHthree}{7}
\soulregister\xspace7
\soulregister\cite7
\soulregister\ref7
\soulregister\pageref7
\soulregister\ie7
\soulregister\eg7
\soulregister\gym7
\soulregister\nn7
\soulregister\aircraft7
\soulregister\newgym7
\soulregister\tunableWeights7
\newcommand{\new}[1]{{#1}}	  % don't highlight
	
\begin{document}

\title{Neuroflight: Next Generation Flight Control Firmware}

\author{
		\IEEEauthorblockN{William Koch, Renato Mancuso, Azer 
			Bestavros}
			\IEEEauthorblockA{Boston University\\
				Boston, MA 02215
					\\\{wfkoch, rmancuso, best\}@bu.edu}
					}

\maketitle
\begin{abstract}
  Quadcopter aircraft are now mainstream and used in a number of
  contexts, including surveillance, aerial mapping and surveying,
  natural disaster recovery, and search and rescue operations. In the
  last decade, seminal results have been accomplished in autonomous
  and semi-autonomous flight. Works in this domain have extensively
  addressed challenges in state estimation, trajectory tracking,
  localization, and the like. Such high-level control objectives rely
  on, and are directly impacted by, the behavior of the inner-most
  control loop, namely attitude control. Remarkably however, attitude
  control has received little-to-no attention in recent years.  Flight
  control implementations predominantly use the classical PID algorithm. PID is easy
  to implement, which resonates well with resource
  constrained quadcopters. Nonetheless, PID has important limitations
  in terms of adaptability. Also, its performance is only as good as
  the tune.

  In this work, we investigate if performing better than PID is
  possible, while still abiding to the typical resource constraints of
  small-scale quadcopters.  For this purpose, we introduce
  Neuroflight, the first open-source neural network-based flight
  controller firmware. We present our toolchain for training a neural
  network in simulation and compiling it to run on embedded
  hardware. We discuss the main challenges faced when jumping from
  simulation to reality along with the proposed solutions. Our
  evaluation shows that the neural network controller can execute at
  over \rMaxLoopHertz on an ARM Cortex-M7 processor. We demonstrate
  via flight tests that a quadcopter running Neuroflight can achieve
  stable flight and execute aerobatic maneuvers.
\end{abstract}

\section{Introduction}
Recently there has been explosive growth in user-level applications
developed for unmanned aerial vehicles~(UAVs). However little
innovation has been made to the UAV's low-level attitude flight
controller which still predominantly uses classic PID
control. Although PID control has proven to be sufficient for a
variety of applications, it falls short in dynamic flight conditions
and environments (\eg in the presence of wind, payload changes and
voltage sags). In these cases, more sophisticated control strategies
are necessary, that are able to adapt and learn.
The use of neural networks~(\nns) for flight control~(\ie neuro-flight control) has been
actively researched for decades to overcome limitations in other control
algorithms such as PID control. However the vast majority of research has
focused on developing autonomous neuro-flight controller autopilots capable of tracking
trajectories~\cite{shepherd2010robust,nicol2008robust,dierks2010output,bagnell2001autonomous,kim2004autonomous,abbeel2007application,hwangbo2017control,dos2012experimental}.
A simple autopilot consists of an \emph{outer loop} and an \emph{inner
loop}. The outer loop is responsible for generating
attitude\footnote{Defined as the orientation of the aircraft in terms
of its angular velocity for each roll, pitch, and yaw axis.}  and
thrust command inputs to follow a specific trajectory. The inner loop
is responsible for maintaining stable flight and for reaching the
attitude set points over time through direct manipulation of the
aircraft's actuators.
Unlike the outer loop, the inner attitude control loop
is {mandatory for both autonomous and manual flight}. Moreover,
the performance of the attitude controller have an important impact on
the ability to reach higher-level control objectives.
This work is the first to explore the adoption of neuro-flight control
as an alternative to PID for inner loop flight control.
Because this work pioneers the adoption of \nn attitude controllers,
we hereby study its performance independently from trajectory
planning. Nonetheless, we acknowledge that studying the implication of
a better attitude controller on higher-level objectives is needed and
planned as future work.

Recently an OpenAI gym~\cite{brockman2016openai}  environment called
\gym~\cite{gymfc} was released.
Via \gym it is possible to train \nns attitude control of a quadcopter
in simulation using reinforcement learning~(RL).  Neuro-flight
controllers trained with Proximal Policy
Optimization~(PPO)~\cite{schulman2017proximal} were shown to exceed
the performance of a PID controller. Nonetheless the attitude
neuro-flight controllers were not validated in the {real world},
thus it remained as an open question if the \nns trained in \gym are
capable of flight.  As such, this work makes the following
contributions:
\begin{enumerate}
\item We introduce Neuroflight, the first open source neuro-flight
controller firmware for multi-rotors and fixed wing aircraft.  The \nn
embedded in Neuroflight replaces attitude control and motor mixing
commonly found in traditional flight control
firmwares~(Section~\ref{sec:overview}).
\item To train neuro-flight controllers capable of stable flight
in the {real world we introduce \newgym}{, a modified
environment addressing several challenges} in making the transition
from simulation to reality~(Section~\ref{sec:challenge}).
\item We propose a toolchain for compiling a trained \nn to run on
embedded hardware. To our knowledge this is the first work that
consolidates a neuro-flight attitude controller on a microcontroller,
rather than a multi-purpose onboard computer, thus allowing deployment
on lightweight micro-UAVs~(Section~\ref{sec:sys}).
\item Lastly, we provide an evaluation showing the
\nn can execute at over \rMaxLoopHertz on an Arm Cortex-M7 processor and
flight tests demonstrate that a quadcopter running \fc can achieve
stable flight and execute aerobatic maneuvers such as rolls, flips,
and the Split-S~(Section~\ref{sec:eval}).  Source code for the project
can be found at
\sourcecode 
and videos of our test flights can be viewed at \videourl.
\end{enumerate}

The goal of this work is to provide the community with a stable
platform to innovate and advance development of neuro-flight control
design for UAVs, and to take a step towards making neuro-flight
controllers mainstream. In the future we hope to establish \nn powered
attitude control as a convenient alternative to classic PID control
for UAVs operating in harsh environments or that require particularly
competitive set point tracking performance~(\eg drone racing).

\section{Background and Related Work}
\label{sec:related}

Since the dawn of fly-by-wire, flight control algorithms have continued to
advance to meet increasing performance
demands~\cite{leith2000survey,hovakimyan2011mathcal,lee2009feedback}.
In recent years a significant amount of research has investigated the use of
\nns for flight control which has advantages over classical control methods
thanks to their ability to \textit{learn} and \textit{plan}. 

Various efforts have demonstrated stable flight of a quadcopter through
mathematical models using neuro-flight controllers to track trajectories. 
Online learning methods
\cite{nicol2008robust,dierks2010output} can learn quadcopter dynamics in
real-time. Yet this requires an initial learning period and flight performance
behavior can be unpredictable for rare occurring events. 
Offline supervised learning \cite{shepherd2010robust} can construct
pre-trained neuro-flight controllers capable of immediate flight.   However realistic data can be expensive to
obtain and inaccuracies from the true aircraft can result in suboptimal control policies. 

RL  is an
alternative to supervised learning for offline learning. It is ideal for
sequential tasks in continuous environments, such as control and thus an attractive
option for training neuro-flight controllers.  RL consists of
an agent (\ie \nn) interacting with an environment to
learn a task. 
At discrete time steps the agent performs an
action (\eg writes control signals to aircraft actuators) in the environment.  
In return the agent receives 
the current state of the environment (obtained from various aircraft sensors which typically becomes the input to
the \nn) and a numerical reward representing the action's performance. The
agent's objective is to maximize its rewards.

Over time there has been a number of successes transferring controllers trained
with RL to a UAVs onboard computer to autonomously track trajectories in the real world.
Flight has been achieved for both
helicopters~\cite{bagnell2001autonomous,kim2004autonomous,abbeel2007application}
and quadcopters~\cite{hwangbo2017control,dos2012experimental}.
Unfortunately none of these works have published any code thereby making it
difficult to reproduce results and to build on top of their research. Furthermore
evaluations are only in respect to the accuracy of position therefore it is
still unknown how well attitude is controlled. 
Of the open source flight control firmwares currently available every single one uses PID control~\cite{ebeid2018survey}.

Koch et. al~\cite{gymfc} proposed an RL environment, \gym, for developing attitude
neuro-flight controllers which exceed accuracy of a PID controller in regards
to angular velocity error.
The \gym environment uses the Gazebo simulator~\cite{koenig2004design},
a high fidelity physics simulator, which contains a
digital replica, or \textit{digital twin}, of the aircraft, fixed about
its center of mass to the simulation world  one meter above the ground
allowing the aircraft to freely rotate in any direction. 
The angular velocity $\Omega(t)=[\Omega_\phi(t),\Omega_\theta(t),\Omega_\psi(t)]$ for each
roll, pitch, and yaw axis of the aircraft is controlled by writing pulse width
modulation~(PWM) values to the aircraft
actuators. 
The agent is trained using episodic tasks~(\ie a task that has a terminal
state). At the beginning of an episodic task a desired angular velocity $\Omega^*(t)$ is randomly sampled. 
The goal of the agent is to achieve this velocity in a finite amount of time starting from  still.
At each time step an action $a(t)=[a_0(t), \dots, a_{N-1}(t)]$ is provided by the
agent where 
$N$ is the number of aircraft actuators to be controlled (\eg $N=4$ for a
quadcopter) and  $a_i(t) \in [1000,2000]$ represents the PWM value.
The agent is returned the state $x(t)=(e(t), \omega(t))$ where 
$e(t)=\Omega^*(t) - \Omega(t)$ is the angular velocity error and
$\omega(t)=[\omega_0(t), \dots, \omega_{N-1}(t)]$ is
the angular velocity of each actuator~(\eg for a quadcopter the RPM of the
motor). Additionally a negative reward $r$ is returned  representing the angular
velocity error. {However evaluations were preformed in simulation thus it was
unknown if neuro-flight controllers trained by \gym could control a quadcopter
in the
real world.}

In this work we pick up where \gym left off. We explain in
Section~\ref{sec:challenge} how without several modifications a \nn trained with
\gym will not be able to achieve stable flight.  
With these modifications addressed in \newgym we were able to generate attitude
neuro-flight controllers capable of high precision flight in the real world.

\section{\fc Overview}
\label{sec:overview}
\fc is a fork of Betaflight version 3.3.3~\cite{betaflight}, a high 
performance flight controller firmware used extensively in
first-person-view~(FPV) multicopter racing.  
Internally Betaflight uses a two-degree-of-freedom PID controller~(not to be
confused with rotational degrees-of-freedom) for attitude 
control and includes other enhancements such 
as gain scheduling for increased stability when battery voltage
is low and  throttle is high.  Betaflight runs on a wide variety of flight 
controller hardware  based on the Arm  Cortex-M family of 
microcontrollers.  Flight control tasks are scheduled using a non-preemptive
cooperative scheduler. 
The main PID controller task consists of multiple
subtasks, including: (1) reading the remote control (RC) command for the desired
angular velocity, (2) reading
and filtering the angular velocity from the onboard
gyroscope sensor, (3) evaluating the PID controller, (4) applying motor mixing to the
PID output to account for asymmetries in the motor locations~(see \cite{gymfc} for
further details on mixing), and (5) writing 
the motor control signals to the electronic speed controller~(ESC). 

\begin{figure*}
 %<left> <lower> <right> <upper>
 \centering
 {\includegraphics[trim=0 280 0
    0,clip,width=0.78\paperwidth]{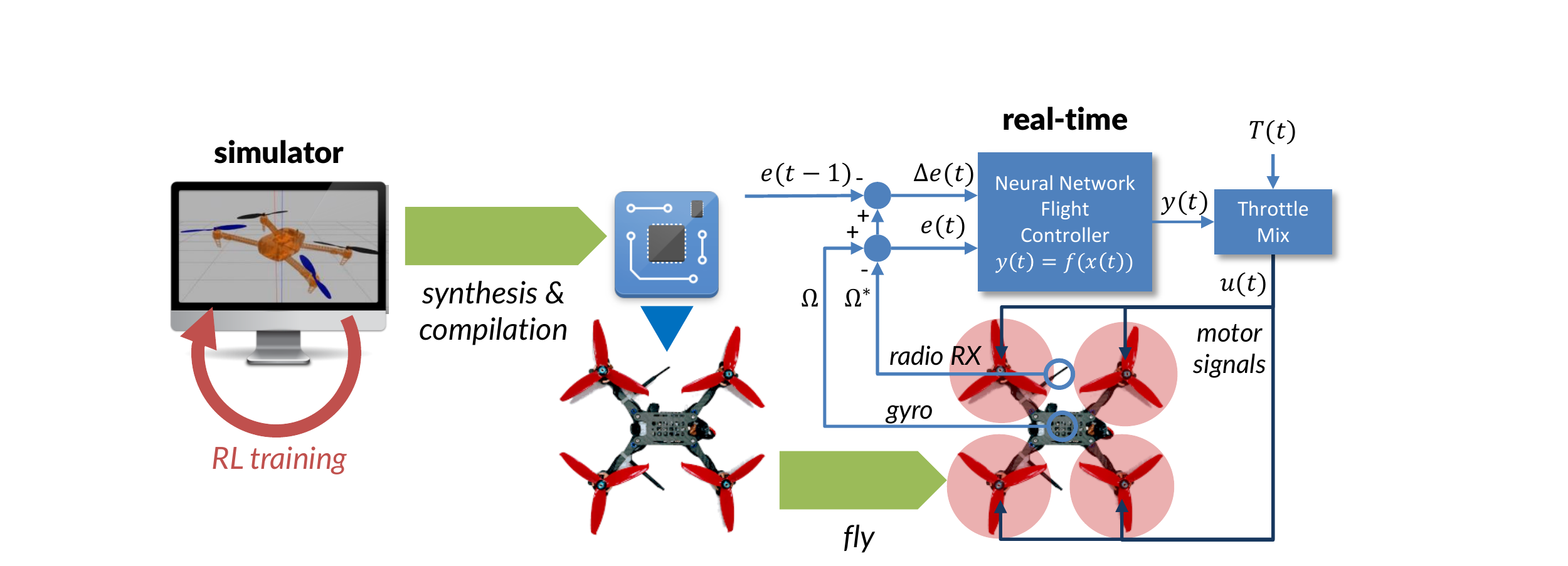}}
 \caption{Overview of the \fc architecture.}
 \label{fig:overview}
\end{figure*}

\fc replaces Betaflight's PID controller task with a neuro-flight controller task.
This task uses a single \nn for attitude control and motor mixing. The
architecture of \fc decouples the \nn from the rest of the firmware
allowing the \nn to be trained and compiled independently.  An
overview of the architecture is illustrated in
Fig.~\ref{fig:overview}.
The compiled \nn is then later linked into \fc to produce a firmware image for the
target flight controller hardware. 

To \fc, the \nn appears to be a generic function $y(t) = f(x(t))$. The
input $x(t)=[e(t), \Delta e(t)]$ where  $\Delta e(t)=e(t) - e(t-1)$. 
The output $y(t)=[y_0, \dots, y_{N-1}]$ where $N$ is the number of
aircraft actuators to be controlled and $y_i \in [0,1]$ is the control
signal representing the percent power to be applied to the $i^{th}$
actuator. This output representation is protocol agnostic and is not
compatible with \nns trained with \gym whose output is the PWM to be
applied to the actuator. PWM is seldomly used in high performance
flight control firmware and has been replaced by digital protocols
such as DShot for improved accuracy and speed~\cite{betaflight}.

At time $t$, the \nn inputs are resolved; $\Omega^*(t)$ is read from the
RX serial port which is either connected to a radio receiver in the case of manual
flight or an onboard companion computer operating as an autopilot in the case of autonomous flight, and $\Omega(t)$ is read from the gyroscope sensor.
 The \nn is then evaluated to obtain the control signal outputs $y(t)$. However the \nn has no concept of thrust
($\mathbf{T}$),
therefore to achieve translational movement the thrust command must be mixed into
the \nn output to produce the final control signal output to the ESC, $u(t)$. 
The logic of throttle mixing is to uniformly apply additional power across all  
actuators proportional to
the available range in the \nn output,
while giving priority to achieving $\Omega^*(t)$. If any output value is over 
saturated (\ie $\exists y_i(t)  : y_i(t) \ge 1$)  no additional throttle will be
added. 
The input throttle value is scaled
depending on the available output range to obtain the actual throttle value:
\begin{equation}
\widehat{\mathbf{T}}(t)= \mathbf{T}(t)\left(1 - \mathrm{max}_i\{y_i(t)\}\right)
\end{equation}
where the function $\mathrm{max}$ returns the max value from the \nn output.
The readjusted throttle value is then proportionally added to each \nn
output to form the final control signal output:
\begin{equation}
    u_i(t)  =
\widehat{\mathbf{T}}(t) + y_i(t).
\end{equation}

\section{\newgym}
\label{sec:challenge}
In this section we discuss the enhancements made to \gym to create \newgym.
These changes primarily consist of a new
state representation and reward system.

\subsection{State Representation}
\label{sec:challenge:motor_dep}
\gym returns the state $(e(t), \omega(t))$ to the agent at each time step.  
However not all UAVs have the sensors to measure motor velocity
$\omega(t)$ as this typically involves digital ESC protocols.  Even in
an aircraft with compatible hardware, including the absolute motor
velocity as an input to the \nn introduces additional challenges. This
is because a \nn trained on absolute RPMs does not easily transfer
from simulation to the real world, unless an accurate propulsion
subsystem model is available for the digital twin.  A mismatch between
the physical propulsion system~(\ie motor/propeller combination) and
the digital twin will result in the inability to achieve stable
flight.
Developing an accurate motor models is time-consuming and
expensive. Specialized equipment is required to capture the relations
between voltage, power consumption, temperature, torque, and thrust.

To address these issues we investigated training using alternative
environment states that do not rely on any specific characteristic of
the motor(s). We posited that reducing the entire state to just
angular velocity errors would carry enough information for the \nn to
achieve stable flight. At the same time, we expected that the
obtained \nn would transfer well to the real aircraft. Thus, our \nn
is trained by replacing $\omega(t)$ with the error differences $\Delta
e(t)$.
To identify the performance impact of this design choice, we trained
two \nns. A first \nn was trained in with $\omega(t)$ in input. Its
behavior was compared to a second \nn trained in an environment that
provides $\Delta e(t)$ instead.
Both \nns were trained with PPO using
hyperparameters from~\cite{gymfc} for 10 million steps.
After training, each \nn was validated against 10 never before seen
random target angular velocities.  Results show the \nn trained in an
environment with $x(t)=(e(t),\Delta e(t))$ experienced on
average \rOmegaDeltaError \textit{less} error with only an increase
of \rOmegaDeltaPower in its control signal outputs.

In RL the interaction between the agent and the environment can be
formally modeled as a Markov Decision Process~(MDP) in which the
probability that the agent transitions to the next state depends on
its current state and action to be taken. The behavior of the agent is
defined by its policy which is essentially a mapping of states to
actions.
There may be multiple different state representations that are able to
map to actions resulting in similar performance.
For instance, it emerged from our experiments that using a history of
errors as input to the \nn also led to satisfactory performance. This
approach has the disadvantage of requiring a \new{state
history table to be maintained}, which ultimately made the approach
less desirable.

The intuition why a state representation comprised of only angular
velocity errors works can be summarized as follows. First, note that a
PD controller (a PID controller with the integrative gain set to zero)
is also a function computed over the angular velocity error. Because
an \nn is essentially a universal approximator, the expectation is
that the \nn would also be able to find a suitable control strategy
based on these same inputs.

Albeit required, however, modifying the environment state alone is not
enough to achieve stable flight. The RL task also needs to be
adjusted.
Training using episodic tasks, in which the aircraft is at rest and
must reach an angular velocity never exposes the agent to scenarios in
which the quadcopter must return to still from some random angular
velocity.
With the new state input consisting of the previous state, this is a
significant difference from \gym which only uses the current state.
For this purpose, a continuous task is constructed to mimic real
flight, continually issuing commands.\footnote{Technically this is
still considered an episodic task since the simulation time is finite.
However in the {real world} flight time is typically finite as
well.} This task randomly samples a command and sets the target
angular velocity to this command for a random amount of time. This
command is then followed by an idle~(\ie $\Omega^*=[0, 0, 0]$) command
to return the aircraft to still for a random amount of time. This is
repeated until a max simulation time is reached.

\subsection{Reward System}
\label{sec:reward}
\new{In the context of RL, reward engineering is the process of designing a
    reward system in order to provide the agent with a numerical  
		signal that they are doing the right
        thing~\cite{dewey2014reinforcement}. Reward engineering is a
        particularly difficult problem. As reward systems increase in
        complexity, they may present unintended side affects resulting in the
        agent behaving in an unexpected manner.}

\newgym reinforces stable flight behavior through our reward
system defined as:
\begin{equation}
r = r_e + r_y + r_\Delta.
\end{equation} 
The agent is penalized
for its angular velocity error, similar to \gym, along each axis with:
\begin{equation}
r_e = -(e_{\phi}^2 + e_{\theta}^2 + e_{\psi}^2).
\end{equation}

However we have
identified the remaining two variables in the reward system as
critical for transferability to the {real world} and achieving
stable flight. Both rewards are a function of the agents control
output. First $r_y$ rewards the agent for minimizing the control
output, and next, $r_\Delta$ rewards the agent for minimizing
oscillations.

The rewards as a function of the control signal are able to aid in the
transferability by compensating for limitations in the training
environment and unmodeled dynamics in the motor model.

\textbf{Minimizing Output Oscillations}
\label{sec:challenge:noise}
In the real world high frequency oscillations in the control output
can damage motors. Rapid switching of the control output causes the
ESC to rapidly change the angular velocity of the motor drawing
excessive current into the motor windings. The increase in current
causes high temperatures which can lead to the insulation of the motor
wires to fail. Once the motor wires are exposed they will produce a
short and ``burn out" the motor.

The reward system used by \gym is strictly a function of the angular
velocity error.  This is inadequate in developing neuro-flight
controllers that can be used in the real world. Essentially this
produces controllers that closely resemble the behavior of an
over-tuned PID controller. The controller is stuck in a state in which
it is always correcting itself, leading to output oscillation.

In order to construct networks that produce smooth control signal
outputs, the control signal output must be introduced into the reward
system. This turned out to be quite challenging.
Ultimately we were able to construct \nns outputting stable control outputs
with the inclusion of the following reward:
\begin{equation}
r_\Delta =
    \beta \sum_{i=0}^{N-1} \text{max}\{ 0, \Delta y_\text{max} -  \left(\Delta
            y_i\right)^2\}.
\end{equation}
This reward is only applied if the absolute angular velocity error for
every axis is less than some threshold~(\ie the error band). This
allows the agent to be signaled by $r_e$ to reach the target without
the influence from this reward.
Maximizing $r_\Delta$ will drive the agent's change in output to zero
when in the error band. To derive $r_\Delta$, the change in the
control output $y_i$ from the previous simulation step is squared to
magnify the effect. This is then subtracted from a constant $\Delta
y_\text{max}$ defining an upper bound for the change in the control
output. The \texttt{max} function then forces a positive reward,
therefore if $(\Delta y_i)^2$ exceeds the limit no reward will be
given. The rewards for each control output $N-1$ are summed and then
scaled by a constant $\beta$, where $\beta > 0$.  Using the same
training and validation procedure previously discussed, we found a \nn
trained in \newgym compared to \gym resulted in
a \ChNoiseYDeltaDecrease decrease in $\Delta y$.

\textbf{Minimizing Control Signal Output Values}
\label{sec:challenge:power}
Recall from Section~\ref{sec:related}, that the \gym environment fixes 
the aircraft to the simulation world about its center of mass, allowing it to
only perform rotational movements. \new{Due to this constraint} the agent can achieve
 $\Omega^*$ with a number of different control signal outputs (\eg when
$\Omega^*=[0,0,0]$ this can be achieved as long as $y_0 \equiv y_1 \equiv y_2
\equiv y_3$).
However this poses a significant problem when transferred to the real world as an aircraft
is not fixed about its center of mass. Any additional power to the motors will result in an
unexpected change in translational movement. This is immediately evident when
arming the quadcopter which should remain idle until RC commands are
received. At idle, the power output (typically ~4\% of the throttle value) must not result in
any translational movement.
Another byproduct of inefficient control signals is a decreased throttle
range~(Section~\ref{sec:overview}). Therefore it
is desirable to have the \nn control signals minimized while still
maintaining the desired angular velocity. 
In order to teach the agent to minimize control outputs we
introduce the reward function:
\begin{equation}
    r_y =\alpha \left( 1-\bar{y} \right)
\end{equation}
providing the agent a positive reward as the output decreases. Since $y_i \leq 1$ 
we first compute the average output $\bar{y}$. Next $1-\bar{y}$ is calculated
as a positive reward for low output usage which is scaled by a constant
$\alpha$, where $\alpha > 0$. \nns trained using this reward experience on average a \ChYDecrease
decrease in their control signal output.

    \new{\textbf{Challenges and Lessons Learned} The fundamental challenge we faced was managing high amplitude oscillations in
the control signal.  
In stochastic continuous control problems it is standard for the network to
output the mean from a Gaussian distribution~\cite{schulman2017proximal,chou2017improving}.
However this poses problems for control tasks with bounded outputs such as
flight control. The typical strategy is to clip the output to the target bounds
yet we have observed this to significantly contribute to oscillations in the control output. 

Through our experience  
we learned that due to the output being stochastic~(which aids in exploration),
the rewards must encapsulate the general trend of the
performance and not necessarily at a specific time~(\eg} \new{the stochastic output
naturally oscillates). 
Additionally we found the reward system must include performance metrics 
other than (but possibly in addition to)
traditional time domain step response characteristics (\eg} \new{overshoot, rise time,
settling time, etc.). Given the agent initially knows nothing, there 
is no step
response to analyze. In future work we will explore the use of goal based
learning in an attempt to develop a hybrid solution in which the agent learns
enough to track a step response, then use traditional metrics for fine tuning.} 

\new{Although our reward system was
sufficient in achieving flight, we believe this is still an open area of
research worth exploring. In addition to aforementioned rewards, we
experimented with several other rewards 
including penalties for over saturation of the
control output (\ie} \new{if the network output exceeded the
clipped region), control output jerk (\ie} \new{change in
acceleration), and the number of oscillations in the output.    
When combining multiple rewards, balancing and tuning the weight can be an
exercise of its own. 
For example if penalizing for number of oscillations or jerk this can lead to
an output that
resembles a low frequency square wave if penalizing the amplitude is not
considered.
We also experimented with positive binary rewards for agent continually making
progress to the desired setpoint however this appear to not provide the agent
with enough information to converge.   
    }

\section{Toolchain}
\label{sec:sys}
\begin{figure}
\centering
{\includegraphics[trim=0 440 520 0,clip,
    width=\columnwidth]{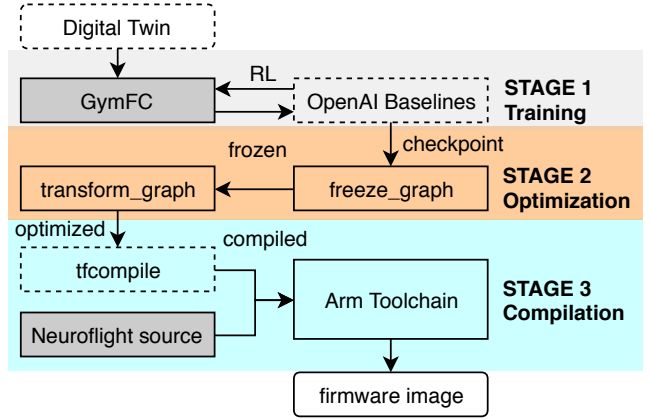}}
\caption{Overview of the \fc toolchain. Our main contributions are in
    the gray
    boxes while boxes with dashed borders indicate minor modifications to
    existing software.}
\label{fig:stages}
\end{figure}

In this section we introduce our toolchain for building the \fc firmware.
\fc is based on the philosophy that each flight control firmware should be
 customized for the target aircraft to achieve maximum performance.
To train a \nn optimal 
attitude control of an aircraft, a digital representation~(\ie a
\textit{digital twin}) of the
aircraft must be constructed to be used in simulation.  
This work begins to address how digital twin fidelity 
affects flight performance, however it is still an open question that we will
address in future work.
The toolchain displayed in {Fig.}~\ref{fig:stages} consists of three stages and takes as input a digital
twin and outputs a \fc firmware unique to the digital twin. 
In the remainder of this section we
will discuss each stage in detail.

\subsection{Training} 

The training stage takes as input a digital twin of an aircraft and outputs a
\nn trained attitude control of the digital twin capable of achieving stable flight in the
real world.
Our toolchain can support any RL library that interfaces
	with  OpenAI environment APIs and allows for the \nn state to be saved  as a
    Tensorflow graph. 
Currently our toolchain uses RL algorithms provided by
OpenAI baselines~\cite{baselines} which has been modified to save the \nn state. 
In Tensorflow, the saved state of a \nn is known as a checkpoint and consists
of three files describing the structure and values in the graph.
Once training has completed, the checkpoint is provides as input to Stage 2: Optimization.

\subsection{Optimization} The optimization stage is an intermediate stage between training and compilation that prepares the \nn graph to be run on hardware. The optimization stage (and compilation stage) require a number of Tensorflow tools which can all be found in the Tensorflow repository~\cite{tensorflow}.  The first step in the optimization stage is to \textit{freeze} the
graph. Freezing the graph accomplishes two tasks: (1) condenses the three checkpoint files into a single Protobuf file by replacing
variables with their equivalent constant values~(\eg numerical weight values)
and (2) extracts the subgraph containing the trained \nn by trimming unused nodes and operations that were only used during training. Freezing is done with Tensorflow's
\texttt{freeze\_graph.py} tool  which takes as input the checkpoint and the
output node of the graph so the tool can identify and extract the subgraph.

Unfortunately the Tensorflow input and output nodes are not documented by RL
libraries (OpenAI baselines~\cite{baselines}, Stable
baselines~\cite{stable-baselines},
TensorForce~\cite{schaarschmidt2017tensorforce})  and in most cases it is not
trivial to identify them. 
We reverse engineered the graph produced by {OpenAI} baselines (specifically the PPO1 implementation) using a
combination of tools and cross referencing with the source code. A Tensorflow graph can
be visually inspected using Tensorflow's Tensorboard tool.
OpenAI baselines does not support Tensorboard thus we created a script
to convert a checkpoint to a Probobuf file and then used Tensorflow's
\texttt{import\_pb\_to\_tensorboard.py}  tool to  view the graph in Tensorboard.
Additionally we used Tensorflow's \texttt{summarize\_graph} tool to 
summarize the inputs and outputs of the graph. 
Ultimately we identified the input node to be ``pi/ob'',  and the output to be 
``pi/pol/final/BiasAdd''.

Once the graph is frozen, it is optimized to run on hardware by running the Tensorflow \texttt{transform\_graph} 
tool. Optimization provided by this tool allows graphs to execute faster and
reduce its overall footprint by further removing unnecessary nodes. 
The optimized frozen ProtoBuf file is provided as input to Stage 3: Compilation.

\subsection{Compilation}
\label{sec:toolchain:compile}
A significant challenge was developing a method to integrate a trained \nn into
\fc to be able to run on the limited resources provided by a 
microcontroller.  The most powerful of the microcontrollers supported
by Betaflight \new{and Neuroflight} consists of 1MB of flash memory\new{, 320KB
of SRAM} and an ARM Cortex-M7
processor with a clock speed of 216MHz~\cite{STM32F745VG}. Recently
there has been an increase in interest for running \nns on embedded
devices but few solutions have been proposed \new{and no standard solution
    exists}.
We found Tensorflow's tool \texttt{tfcompile} to work
best for our toolchain. \texttt{tfcompile} provides 
ahead-of-time (AOT) compilation of Tensorflow graphs into executable code 
primarily motivated as a method to execute graphs on mobile devices.  Normally executing 
graphs requires the Tensorflow runtime  which is far too heavy  
for a microcontroller.  Compiling graphs using \texttt{tfcompile} does 
not use the Tensoflow runtime which results in a self contained executable and
a reduced footprint. 

Tensorflow uses the
Bazel~\cite{bazel} build system and  expects you
will be using the \texttt{tfcompile} Bazel macro in your project. \fc on
the other hand is using \texttt{make} with the GNU Arm Embedded Toolchain. Thus
it was necessary for us to integrate \texttt{tfcompile} into the toolchain by calling the \texttt{tfcompile} binary directly.
When invoked, an object file representing the compiled
graph  and an accompanying header file is produced. Examining the header file we identified three additional Tensorflow dependencies that must be included
in \fc~(typically this is automatically included if using the Bazel build system): the AOT runtime 
(\texttt{runtime.o}), an interface to run the compiled functions
(\texttt{xla\_compiled\_cpu\_function.o}), and running options
(\texttt{executable\_run\_options.o}) for a total of \depsize. 
In 
Section~\ref{sec:eval} we will analyze the size of the generated object file
for the specific neuro-flight controller.

To perform fast floating point calculations \fc must be compiled with
ARM's hard-float application binary interface~(ABI). Betaflight core,
inherited by
\fc already defines the proper compilation flags in the Makefile however it is
required that the entire firmware must be compiled with the same ABI
meaning the Tensorflow graph must also be compiled with the same ABI.
Yet \texttt{tfcompile} does not currently allow for setting arbitrary
compilation flags which required us to modify the code.  Under the
hood, \texttt{tfcompile} uses the LLVM backend for code generation. We
were able to enable hard floating points through the ABIType attribute
in the \texttt{llvm::TargetOptions} class.

\section{Evaluation}
\label{sec:eval}
\begin{figure*}
  \begin{subfigure}[b]{0.5\textwidth}
    \includegraphics[width=\textwidth]{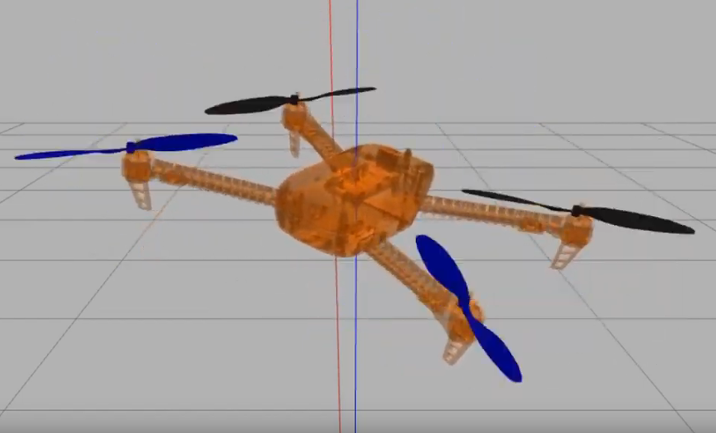}
    \caption{Screenshot of the Iris quadcopter flying in simulation.}
    \label{fig:iris_sim}
  \end{subfigure}
  ~
  \begin{subfigure}[b]{0.5\textwidth}
    \includegraphics[trim=0 0 0
      0,clip,width=\textwidth]{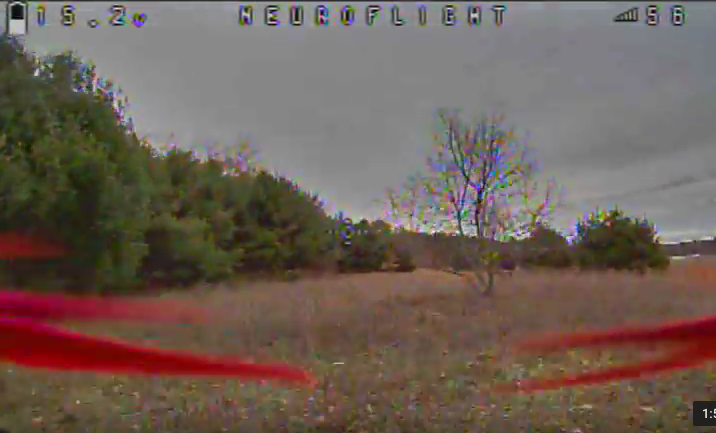}
    \caption{Still frame of the FPV video footage acquired during a test
        flight.}
    \label{fig:nf1_fpv}
  \end{subfigure}
  \caption{Flight in simulation (left) and in the real world (right).}
\end{figure*}

In this section we evaluate \fc controlling a high performance \new{custom} FPV
racing quadcopter \new{named \aircraft}. First and foremost, we show that it is capable of
maintaining stable flight. Additionally, we demonstrate that the
synthesized \nn controller is also able to stabilize the aircraft even
when executing advanced aerobatic maneuvers.  Images of \new{\aircraft}
and its entire build log have been published to
RotorBuilds~\cite{rotorbuild}.

\subsection{Firmware Construction} 
\label{sec:eval:firmware}
We used the Iris quadcopter model included with the Gazebo
simulator \new{(which is also used by \gym)} with modifications to the motor model \new{to more accurately reflect \aircraft} for our digital twin.
The digital twin motor model used by Gazebo is quite simple. Each
control signal is multiplied by a maximum rotor
velocity constant to derive the target rotor velocity while each rotor is
associated with a PID
controller to achieve this target rotor velocity. 
We obtained an estimated maximum 33,422 RPMs for our propulsion system
from Miniquad Test Bench~\cite{miniquad} to update the maximum rotor
velocity constant. We also modified the rotor PID controller (P=0.01,
I=1.0) to achieve a similar throttle ramp.

\begin{figure}
\centering
\begin{subfigure}[b]{0.2\textwidth}
    {\includegraphics[trim=20 0 53 30,clip,
        width=\textwidth]{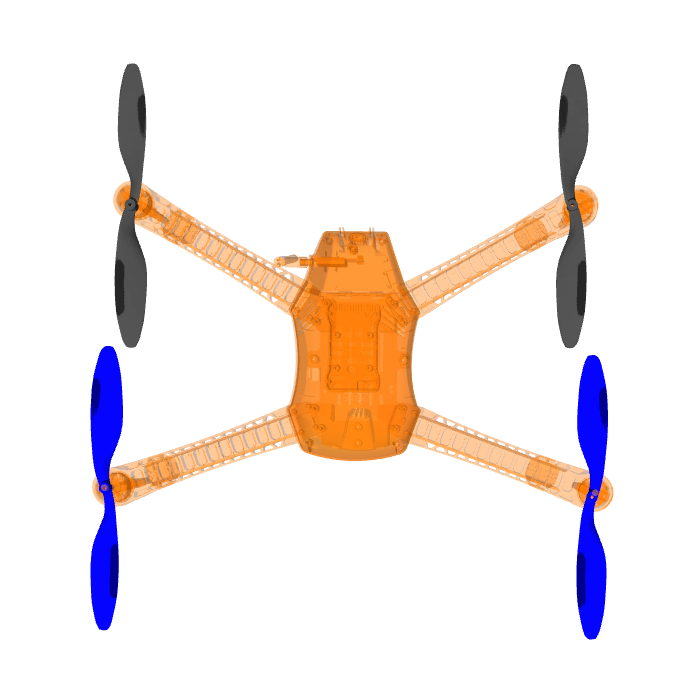}}
    \caption{Iris }
    \label{fig:iris}
\end{subfigure}
~
\begin{subfigure}[b]{0.2\textwidth}
    {\includegraphics[trim=20 0 53 30,clip, width=\textwidth]{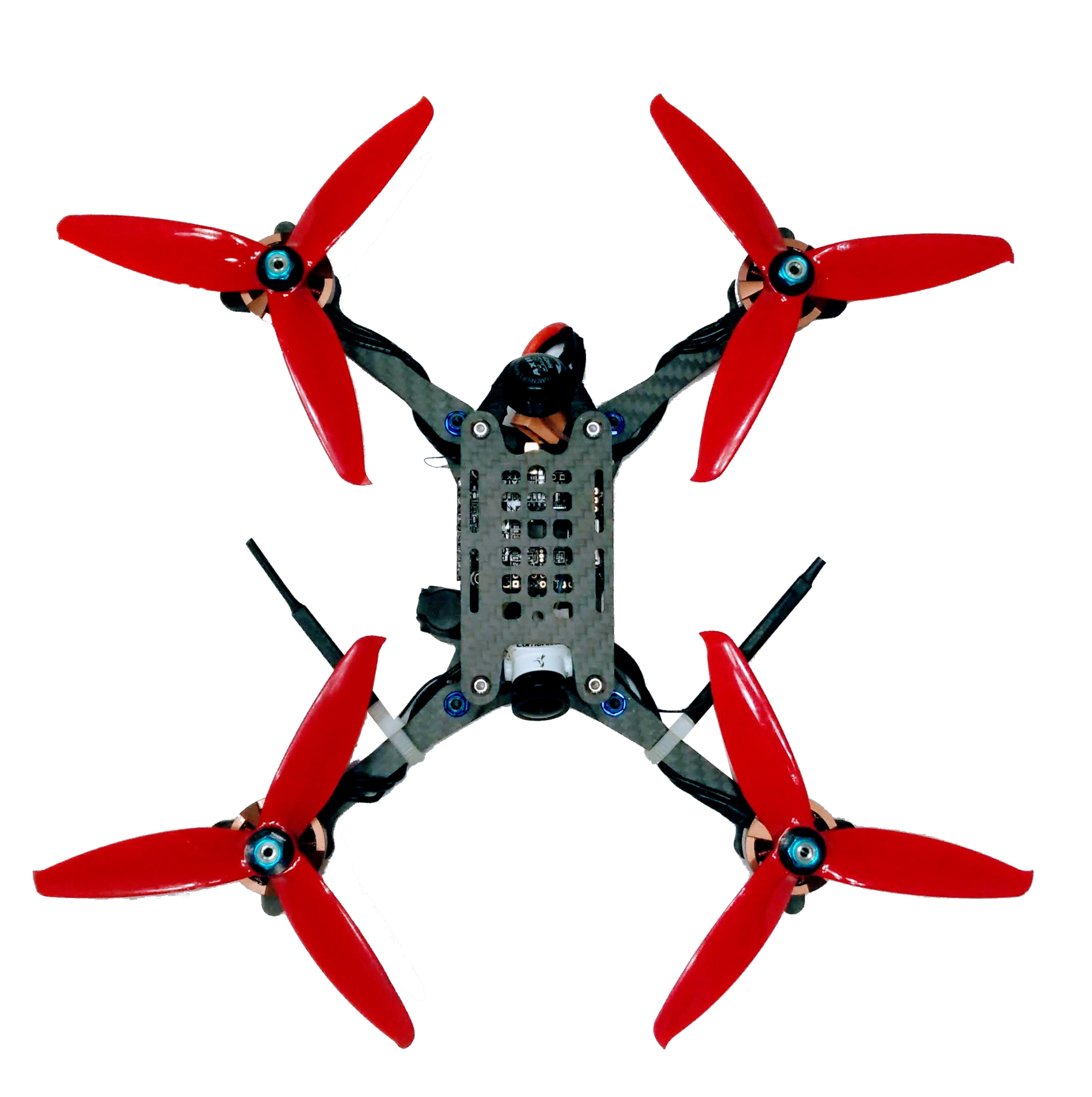}}
    \caption{\aircraft}
    \label{fig:quad}
\end{subfigure}
\caption{{Iris simulated quadcopter compared to the \aircraft}{ real
        quadcopter.}} 
\label{fig:quadcompare}
\end{figure}

{\aircraft}{ is in stark contrast with the Iris quadcopter
  model used by \gym}{ which is advertised for autonomous flight
  and imaging~\cite{iris}. We have provided a visual comparison in
  Fig.~\ref{fig:quadcompare} and a comparison between the aircraft
  specifications in Table~\ref{tbl:specs}.  In this table, weight
  includes the battery, while the wheelbase is the motor to motor
  diagonal distance. Propeller specifications are in the format
  ``LL:PPxB" where LL is the propeller length in inches, PP is the
  pitch in inches and B is the number of blades. Brushless motor sizes
  are in the format ``WWxHH" where WW and HH is the stator width and
  height respectively. The motors $K_v$ value is the motor velocity
  constant and is defined as the inverse of the motors back-EMF
  constant which roughly indicates the RPMs per volt on an unloaded
  motor~\cite{learnrc}. Flight controllers are classified by the
  version of the embedded ARM Cortex-M processor prefixed by the
  letter `F' (\eg}{ F4 flight controller uses an ARM Cortex-M4).}

\begin{table}[]
    \centering
    \begin{tabular}{l|ll}
                          & Iris                    & \aircraft
        \\ \hline
		Weight			  & 1282g       & 432g
		\\
		Wheelbase		  & 550mm  					& 212mm
		\\ \hline
        Propeller         & 10:47x2                  & 51:52x3
        \\
        Motor             & 28x30 850$K_v$              & 22x04 2522$K_v$
        \\
        Battery           & 3-cell 3.5Ah LiPo & 4-cell 1.5Ah LiPo
        \\
        Flight Controller & F4                      & F7                     
    \end{tabular}
\caption{{Comparison between Iris and \aircraft}{ specifications.} }
\label{tbl:specs}
\end{table}

Our \nn architecture consists of \new{6 inputs, 4 outputs,} 2 hidden layers with 32 nodes each
using hyperbolic tangent activation functions \new{ resulting in a total of
    \tunableWeights} \new{ tunable weights}. \new{The network outputs the mean of a Gaussian distribution with a variable standard deviation as defined by PPO for continuous domains~\cite{schulman2017proximal}.}
Training was performed with the OpenAI Baseline version 0.1.4 implementation of PPO1
due to its previous success in~\cite{gymfc} \new{which showed PPO to out perform Deep
Deterministic Policy Gradient~(DDPG)~\cite{lillicrap2015continuous}, and Trust Region Policy
Optimization~(TRPO)~\cite{schulman2015trust} in regards to attitude control
in simulation}. A picture of the quadcopter during  
trained in \newgym can be seen in Fig.~\ref{fig:iris_sim}.

The reward system hyperparameters used were $\alpha=300$, $\beta=0.5$, and
$\Delta y_{max}=100^2$ \new{and the PPO hyperparameters used are reported in
    Table~\ref{tab:ppo}. The reward hyperparameter $\Delta y_{max}$ is defined
    as the maximum delta in the output we are willing to accept, while $\alpha$
    and $\beta$ were found through experimentation to find the desired balance between
    minimizing the output and minimizing the output oscillations.  The discount
    and Generalized Advantage Estimate~(GAE) parameters were taken from
\cite{schulman2017proximal} while the remaining parameters were found using
random search. The agent was particularity sensitive to the selection
of the horizon and minibatch size. To account for sensor noise in the real
world we added noise to the angular velocity measurements which was sampled from a
Gaussian distribution with $\mu=0$ and $\sigma=5$. The standard deviation was
obtained by
incrementing $\sigma$ until it began to impact the controllers ability to track
the set point. We observed 
this to reduce motor oscillations in the real world.}

\begin{table}[]
    \centering
\begin{tabular}{l|c}
Hyperparameter            & Value          \\ \hline
Horizon (T)               & \rPpoHorizon  \\
Adam stepsize             & \rPpoStepsize  \\
Num. epochs               & \rPpoEpochs    \\
Minibatch size            & \rPpoMinibatch \\
Discount ($\gamma$)       & \rPpoDiscount  \\
GAE parameter ($\lambda$) & \rPpoGae      
\end{tabular}
\caption{\new{PPO Hyperparameters where $\rho$ is linearly annealed over the
course of training from 1 to 0.}}
\label{tab:ppo}
\end{table}

\begin{figure}
\centering
{\includegraphics[trim=0 0 0
    0,clip,width=1.0\columnwidth]{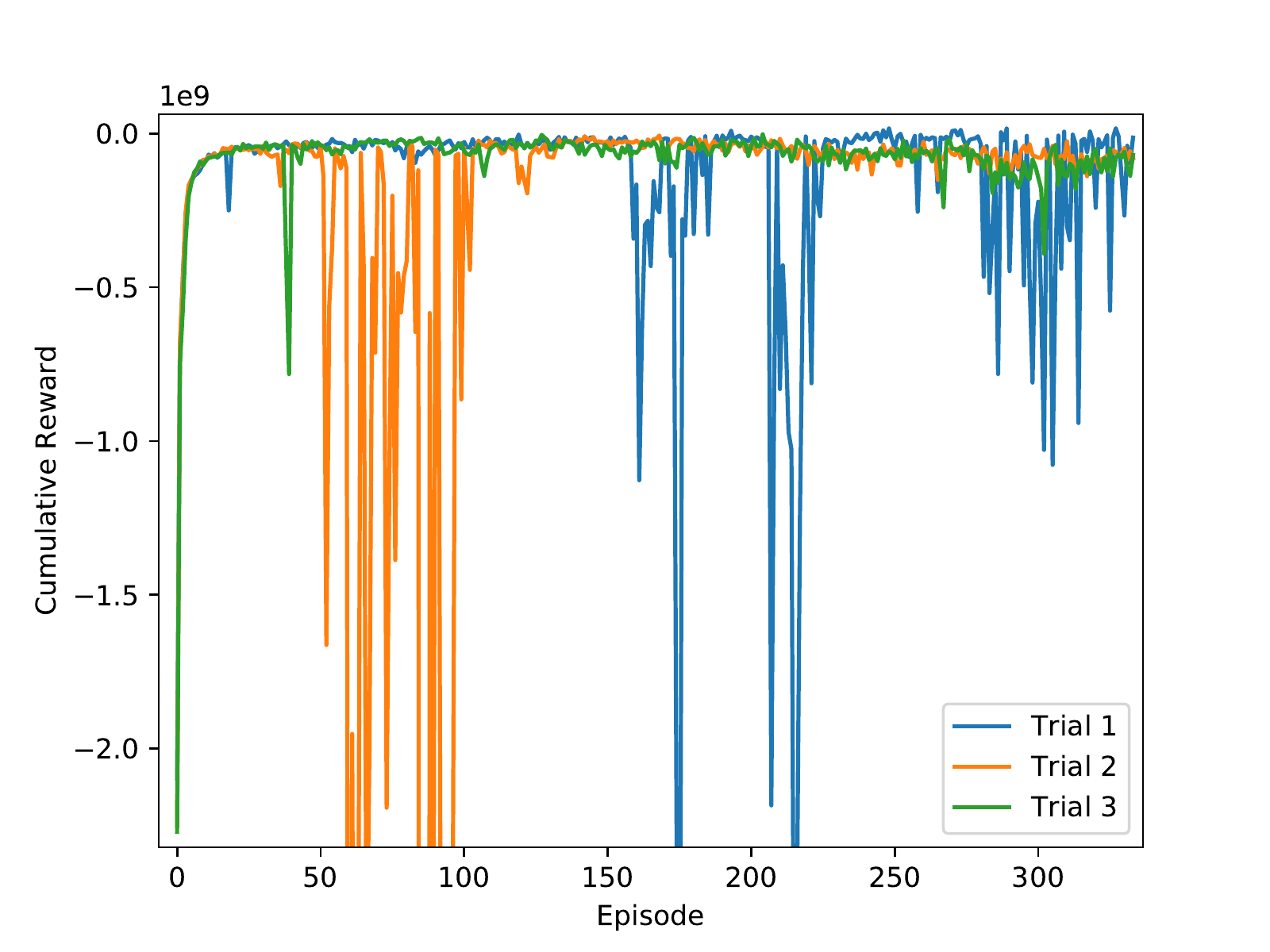}}
\caption{Cumulative rewards for each training episode.}
\label{fig:rewards}
\end{figure}

Each training task/episode ran for 30
seconds in simulation. The simulator is configured to take simulation steps
every 1ms which results in a total of 30,000 simulation steps per episode.
\new{Training ran for a total of 10 million time steps (333 episodes) on a desktop computer
    running Ubuntu 16.04 with an eight-core i7-7700 CPU and an NVIDIA GeForce GT
    730 graphics card which took approximately 11 hours. However training
    converged much earlier at around 1 million time steps (33 episodes) in just
    over an hour~(Fig.~\ref{fig:rewards}).}
We trained a total of three \nns which each used a different random seed for
the RL training algorithm and selected the \nn that received the highest
cumulative reward to use in \fc. Fig.~\ref{fig:rewards} shows a plot of the cumulative
rewards of each training episode for each of the \nns. 
The
plot illustrates how drastic training episodes can  vary simply due to the use
of a different seed.

The optimization stage reduced the frozen Tensorflow graph of the best
performing \nn by
\rOptGraphDecrease to a size of \rGraphSize.
The graph was compiled with Tensorflow version 1.8.0-rc1 and the firmware was compiled for the MATEKF722 target corresponding to the
manufacturer and model of our flight controller MATEKSYS Flight Controller
F722-STD. \new{Our flight controller uses the STM32F722RET6 microcontroller
with 512KB flash memory, and 256KB of SRAM.}

\new{
We inspected the \texttt{.text}, \texttt{.data} and \texttt{.bss} section
    headers of the firmware's ELF file to derive a lower bound of the memory
    utilization. These sections totalled 380KB, resulting in at least 74\% utilization
    of the flash memory. Graph optimization accounted for a reduction
    of 280B, all of which  was reduced from the \texttt{.text} section.
    Although in terms of memory utilization the  optimization stage was not
    necessary, this however will be more important for
    larger networks in the future.
    Comparing this to the parent project, Betaflight's sections totalled 375KB.

    Using Tensorflow's benchmarking tool we performed one million evaluations of the
    graph with and without optimization and found the optimization processes
    to reduce execution time on average by $1.1 \mu s$.

}

\subsection{Timing Analysis}
\label{sec:timing}

Running a flight control task with a fast control rate allows for the use of a
high speed ESC protocol, reducing write latency to the motors and thus resulting in higher precision flight. 
Therefore it is critical to analyze the
execution time of the neuro-flight control task so the optimal
control rate of the task can be determined. Once this is identified it can be
used to select which 
ESC protocol will provide the best performance. 
We collect timing data for \fc and compare
this to its parent project Betaflight. Times are taken for when the quadcopter is disarmed and also armed under
load for the control algorithm  (\ie evaluation of the \nn and PID
equation) and also the entire flight control task which in addition to the
control algorithm includes reading the gryo, reading the RC commands and
writing to the motors.
\begin{table}[]
    \centering
\begin{tabular}{ll|c|c|c|}
\cline{3-5}
                                                &             & WCET~($\mu s$)
& BCET~($\mu s$)                     & Var. Window (\%)         \\ \hline
\multicolumn{1}{|l|}{\multirow{2}{*}{Disarmed}} & NF & \rTimeNfAlgDisarmedWCET & \rTimeNfAlgDisarmedBCET & \rTimeNfAlgDisarmedP \\ \cline{2-5} 
\multicolumn{1}{|l|}{}                          & BF  & \rTimeBfAlgDisarmedWCET  & \rTimeBfAlgDisarmedBCET & \rTimeBfAlgDisarmedP \\ \hline
\multicolumn{1}{|l|}{\multirow{2}{*}{Armed}}    & NF & \rTimeNfAlgArmedWCET    & \rTimeNfAlgArmedBCET    & \rTimeNfAlgArmedP    \\ \cline{2-5} 
\multicolumn{1}{|l|}{}                          & BF  & \rTimeBfAlgArmedWCET    & \rTimeBfAlgArmedBCET    & \rTimeBfAlgArmedP    \\ \hline
\end{tabular}
\caption{Timing analysis of the control algorithm used in Neuroflight~(NF) and
Betaflight~(BF).}
\label{table:timing_alg}
\end{table}

We instrumented the firmware to calculate the timing measurement and
wrote the results to an unused serial port on the flight control
board.  Connecting to the serial port on the flight control board via
an FTDI adapter we are able to log the data on an external PC running
\texttt{minicom}. We recorded 5,000 measurements and report the
worst-case execution time~(WCET), best-case execution time~(BCET) and
the variability window in Table~\ref{table:timing_alg} for the control
algorithm and Table~\ref{table:timing_loop} for the control task.  The
variability window is calculated as the difference between the WCET
and BCET, normalized by the WCET, \ie
$(\text{WCET}-\text{BCET})/\text{WCET}$. This provides indication of
how predicable is the execution of the flight control logic, as it
embeds information about the relative fluctuation of execution
times. Two remarks are important with respect to the results in
Table~\ref{table:timing_alg}. First, the \nn traversal compared to PID
is about 14x slower (armed case), albeit the predictability of the
controller increases. It is important to remember that, while
executing the PID is much simpler than evaluating an \nn, our approach
allows removing additional logic that is required by a PID, such as motor mixing. Thus, a more meaningful comparison
needs to be performed by looking at the overall WCET and
predictability of the whole flight control task, which we carry out in
Table~\ref{table:timing_loop}. Second, because \nn evaluation always
involve the same exact steps, an improvement in terms of
predictability can be observed under Neuroflight.

\begin{table}[]
    \centering
\begin{tabular}{ll|c|c|c|}
\cline{3-5}
                                                &             & WCET~$(\mu s)$
& BCET~$(\mu s)$           & Var. Window (\%)           \\ \hline
\multicolumn{1}{|l|}{\multirow{2}{*}{Disarmed}} & NF & \rTimeNfLoopDisarmedWCET & \rTimeNfLoopDisarmedBCET & \rTimeNfLoopDisarmedP \\ \cline{2-5} 
\multicolumn{1}{|l|}{}                          & BF  & \rTimeBfLoopDisarmedWCET & \rTimeBfLoopDisarmedBCET & \rTimeBfLoopDisarmedP \\ \hline
\multicolumn{1}{|l|}{\multirow{2}{*}{Armed}}    & NF & \rTimeNfLoopArmedWCET    & \rTimeNfLoopArmedBCET    & \rTimeNfLoopArmedP    \\ \cline{2-5} 
\multicolumn{1}{|l|}{}                          & BF  & \rTimeBfLoopArmedWCET    & \rTimeBfLoopArmedBCET    & \rTimeBfLoopArmedP    \\ \hline
\end{tabular}
\caption{Timing analysis of the flight control task used in Neuroflight~(NF) and
Betaflight~(BF).}
\label{table:timing_loop}
\end{table}

The timing analysis reported in Table~\ref{table:timing_loop} reveals
that the neuro-flight control task has a WCET of \rTimeNfLoopArmedWCET
$\mu s$ which would allow for a max execution rate of \rMaxNNTaskFreq.
However in \fc (and in Betaflight), the flight control task frequency
must be a division of the gyro update frequency, thus with
\rMaxGyroHertz gyro update and a denominator of \rLoopDenom, the
neuro-flight control task can be configured to execute at
\rMaxLoopHertz. To put this into perspective this is \rFasterThanPX
times faster\footnote{According to the default loop rate of 250Hz.}
than the popular PX4 firmware~\cite{meier2015px4}.

Furthermore this control rate is \rFastThanPWM times faster than the traditional PWM ESC
protocol used by commercial quadcopters (50Hz \cite{abdulrahim2019defining}) 
thereby allowing us to configure \fc  to use the ESC protocol DShot600 which has
a max frequency of 37.5kHz ~\cite{looptime}.   

\begin{figure*}
\centering
{\includegraphics[trim=50 0 70
    0,clip,width=\textwidth]{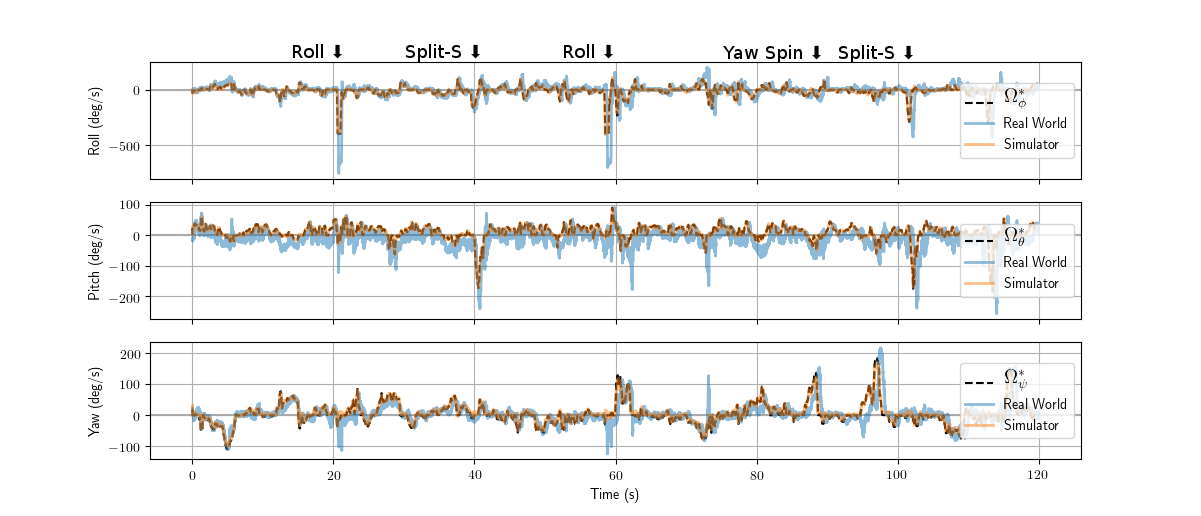}}
\caption{Flight test log demonstrating \fc tracking a desired angular velocity
    in the {real world} compared to
    in simulation. Maneuvers during this flight are annotated. }
\label{fig:flight}
\end{figure*}

Given the simplicity of the PID algorithm it 
came as no surprise that the Betaflight flight control task is faster, yet this is only by a
factor of \rSlowerThanBF when armed. As we can see comparing  
Table~\ref{table:timing_alg} to Table~\ref{table:timing_loop} the additional
subprocesses tasks  are the bottleneck of the
Betaflight flight control ask. 
However referring to the variability window, the \fc control algorithm
and control task are far more stable than Betaflight. The Betaflight
flight control task exhibits little predictability when armed.

Recent research has shown there is no measurable improvements for control task
loop rates that are faster than 4kHz~\cite{abdulrahim2019defining}.
Our timing analysis has shown that \fc is close of this goal.  To
reach this goal there are three approaches we can take: (1) Support
future microcontrollers with faster processor speeds, (2) experiment
with different \nn architectures to reduce the number of arithmetic
operations and thus reduce the computational time to execute the \nn,
and (3) optimize the flight control sub tasks to reduce the flight
control task's WCET and variability window. In future work we
immediately plan to explore (2) and (3), results obtained in these
directions would not depend on the specific hardware used in the final
assembly.

\begin{figure}
\centering
\begin{subfigure}[b]{0.9\columnwidth}
    {\includegraphics[trim=0 0 20 20,clip, width=\textwidth]{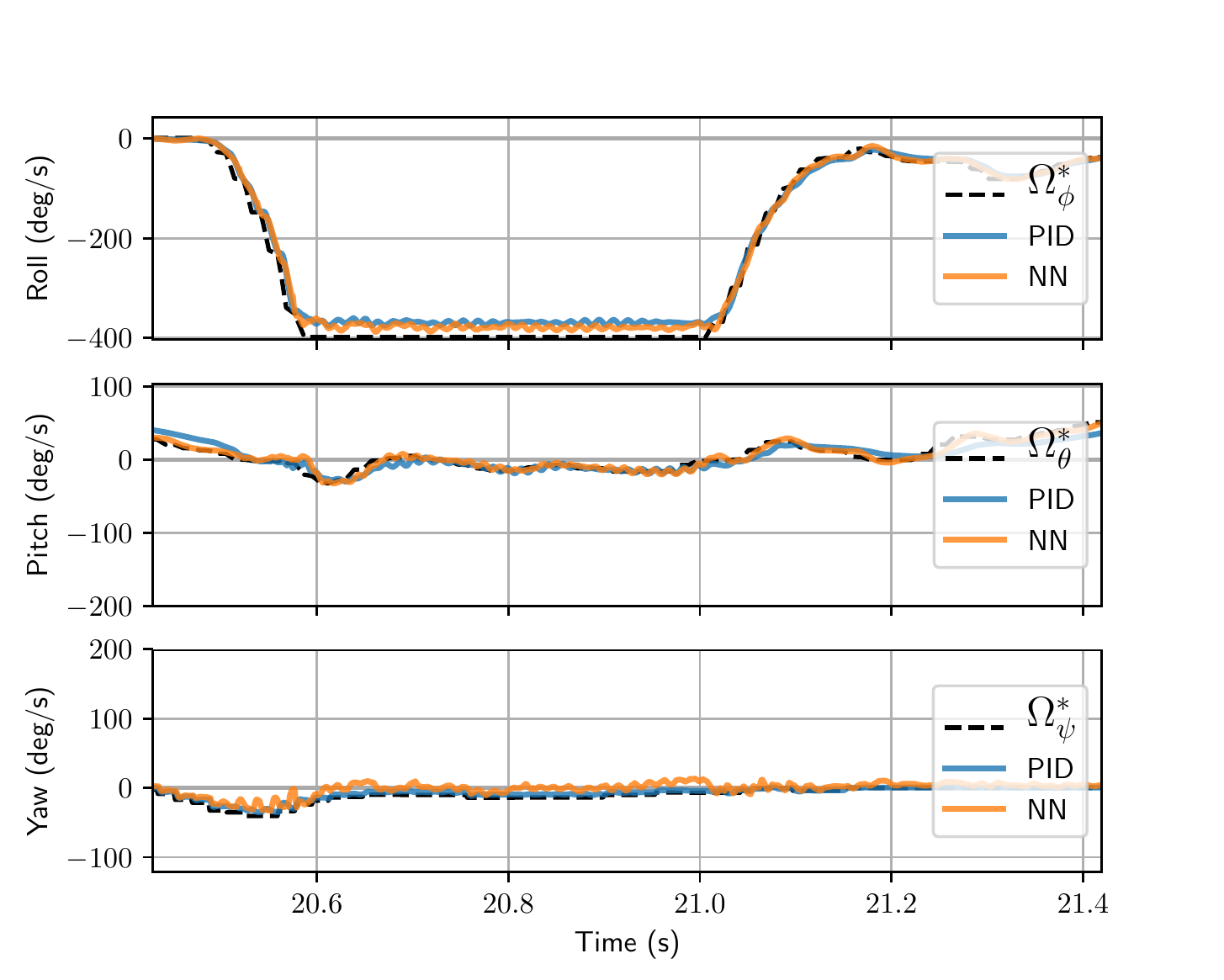}}
    \caption{\new{Roll}}
    \label{fig:roll}
\end{subfigure}
\\
\begin{subfigure}[b]{0.9\columnwidth}
    {\includegraphics[trim=0 0 20 20,clip,
        width=\textwidth]{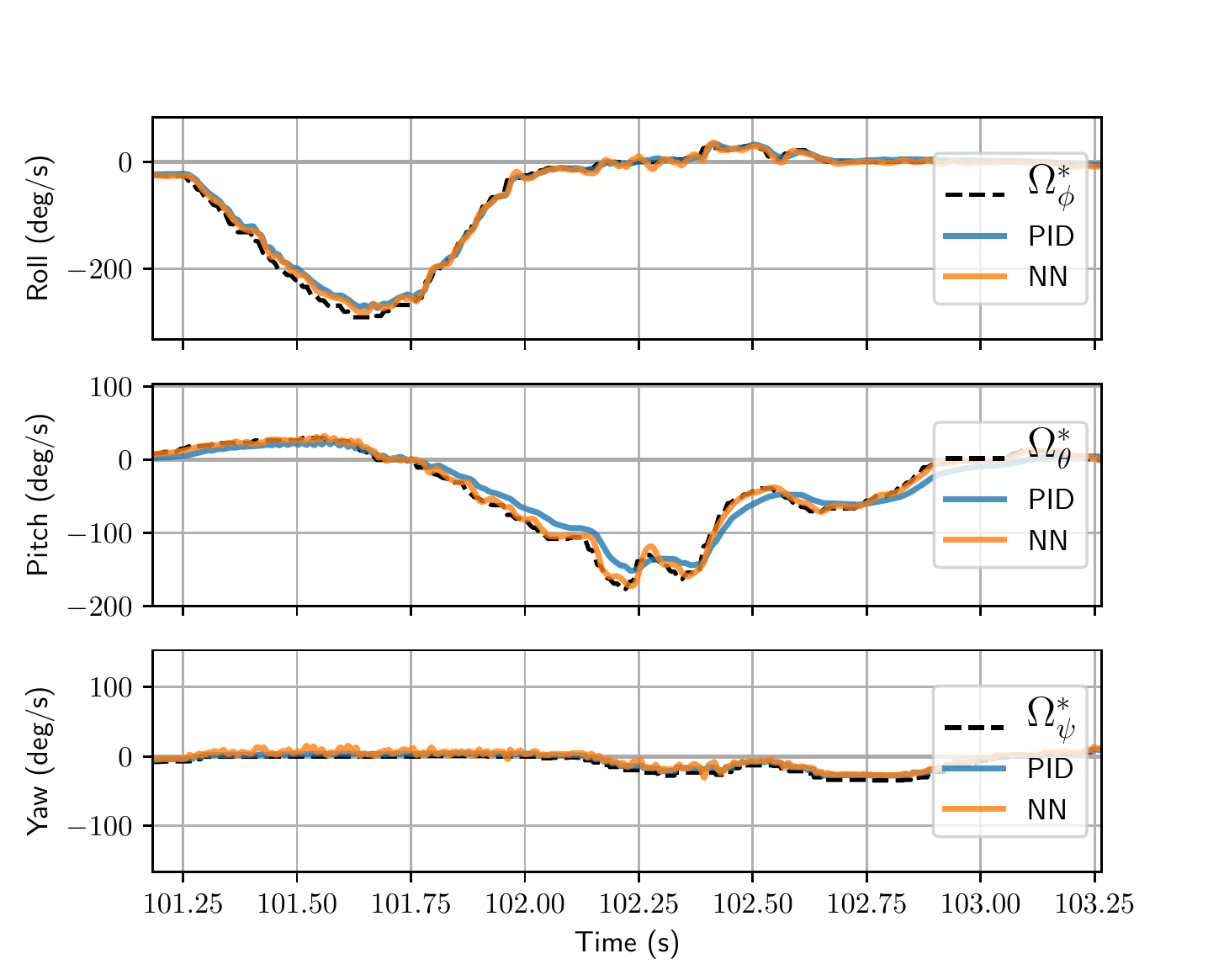}}
    \caption{\new{Split-S} }
    \label{fig:splits}
\end{subfigure}
\caption{\new{Performance comparison of the \nn} \new{ controller versus a PID controller tracking a desired angular velocity
    in  simulation to execute the Split-S and roll aerobatic maneuvers.}} 
\label{fig:zoom}
\end{figure}

\subsection{Power Analysis}
\label{sec:power}
\new{The flight controller affects power consumption directly and indirectly.
    The direct power draw is a result of the execution of the control
    algorithm/task, while the indirect power draw is due to the generated control
    signals which determines the amount of power the ESC will draw.} 
    
\new{As a first attempt to understand and compare the power consumption of a \nn} \new{based controller  to a standard PID controller, we performed a
    static power analysis. For \aircraft} \new{running Neuroflight, we connected a multimeter
    inline with the battery power supply to measure the current
    draw and report the measurements for both when the quadcopter is
    disarmed~(direct power consumption) 
    and armed idling~(indirect power consumption), similarly done to our timing
    analysis. We then take the same measurements
    for the \aircraft} \new{running Betaflight~(PID control). Results reported in
    Table~\ref{table:power} show there is no change using the \nn} \new{based
    controller in regards to direct power draw of the control algorithm. This result
    was expected as the flight control firmware does not execute sleep
    instructions. However for the indirect power draw, there is a measurable
    70mA~(approximately 11\%) increase in
    current draw for the \nn} \new{controller. 
It is important to remember this particular \nn} \new{controller has been trained to
optimize its ability to track a desired angular velocity. Thus the increase in
current draw does not come as a surprise as the control signals will be
required to switch
quickly to maintain the set point which results in increased current draw.}

\new{An advantage a \nn} \new{
controller has over a traditional PID controller is that it has the ability to
optimize its performance based on a number of conditions and characteristics, such as power
consumption.  In the future we will investigate alternative optimization goals
for the controller and instrument \aircraft} \new{ with sensors to record power
consumption in flight to perform a thorough power analysis.
}

\begin{table}[]
    \centering
\begin{tabular}{ll|c|c|c|}
\cline{3-5}
                                                &    & Voltage (V)   & Current (A)     & Power (W)        \\ \hline
\multicolumn{1}{|l|}{\multirow{2}{*}{Disarmed}} & NF & \powerVoltage & \ampsNFdisarmed & \powerNFdisarmed \\ \cline{2-5} 
\multicolumn{1}{|l|}{}                          & BF & \powerVoltage & \ampsBFdisarmed & \powerBFdisarmed \\ \hline
\multicolumn{1}{|l|}{\multirow{2}{*}{Armed}}    & NF & \powerVoltage & \ampsNFarmed    & \powerNFarmed    \\ \cline{2-5} 
\multicolumn{1}{|l|}{}                          & BF & \powerVoltage & \ampsBFarmed    & \powerBFarmed    \\ \hline
\end{tabular}
\caption{\new{Power analysis of  Neuroflight (NF) compared to Betaflight (BF).}}
\label{table:power}
\end{table}

\subsection{Flight Evaluation}
\label{sec:flighteval}

To test the performance of \fc we had an experienced drone racing
pilot conduct test flights for us. The FPV videos of the test flights can be viewed
at \videourl. A still image extracted from the
FPV video feed shows the view point of the pilot of one of the test
flights can be seen in Fig.~\ref{fig:nf1_fpv}. In FPV flying the
aircraft has a camera which transmits the analog video feed back to
the pilot who is wearing goggles with a monitor connected to a video
receiver. This allows the pilot to control the aircraft from the
perspective of the aircraft.

\fc supports real-time logging
during flight allowing us to collect gyro and RC command data to analyze how
well the neuro-flight controller is able to track the desired angular velocity. 
We asked the pilot to fly a mix of basic maneuvers such as loops and figure
eights and advanced maneuvers such as rolls, flips, dives and the Split-S. To execute
a Split-S the pilot inverts the quadcopter and descends in a half loop dive, exiting the loop so they are flying in the opposite horizontal direction. 
Once we collected the flight logs we played the desired angular rates back to
the \nn in the \newgym environment to evaluate the performance in simulation.
Comparison between the simulated and {real world} performance is illustrated in
{Fig.}~\ref{fig:flight} while specific maneuvers that occur during this test
flight are annotated. 
\new{
We computed various control measures of the flight performance including the
Mean Absolute Error~(MAE), 
and 
Mean Squared Error~(MSE),
as well as the descrete form of the
Integral Absolute Error~(IAE), 
Integral Squared Error~(ISE), 
Integral Time-weighted Absolute Error~(ITAE),
and
Integral Time-weighted Squared Error~(ITSE).
These values are reported 
 in Table~\ref{tab:real_flight} for the real flight
and in Table~\ref{tab:flight_metrics} for the simulated flight .
}
\begin{table}[]
    \centering
\begin{tabular}{l|ccc|c|}
\cline{2-5}
                             & \multicolumn{4}{c|}{NN Controller (PPO)}                                                \\ \hline
\multicolumn{1}{|l|}{Metric} & Roll ($\phi$) & Pitch($\theta$) & Yaw ($\psi$)  & \cellcolor[HTML]{DAE8FC}Average       \\ \hline
\multicolumn{1}{|l|}{MAE}    & 15         & 21           & 11         & \cellcolor[HTML]{DAE8FC}15         \\
\multicolumn{1}{|l|}{MSE}    & 1,010      & 595          & 282        & \cellcolor[HTML]{DAE8FC}629        \\
\multicolumn{1}{|l|}{IAE}    & 15,094     & 20,946       & 11,203     & \cellcolor[HTML]{DAE8FC}15,748     \\
\multicolumn{1}{|l|}{ISE}    & 1,005,923  & 592,318      & 281,548    & \cellcolor[HTML]{DAE8FC}626,596    \\
\multicolumn{1}{|l|}{ITAE}   & 911,269    & 1,246,625    & 678,602    & \cellcolor[HTML]{DAE8FC}945,499    \\
\multicolumn{1}{|l|}{ITSE}   & 52,904,984 & 36,036,125   & 18,023,064 & \cellcolor[HTML]{DAE8FC}35,654,724 \\ \hline
\end{tabular}
\caption{\new{Performance metrics of NN controller from flight in the real
world. Metric is reported for each individual axis, along with the average.
Lower values are better.}}
\label{tab:real_flight}
\end{table}

\begin{table*}[h]

    \centering
\begin{tabular}{l|ccc|c|ccc|c|}
\cline{2-9}
                             & \multicolumn{4}{c|}{NN Controller (PPO)}                                              & \multicolumn{4}{c|}{PID}                                                              \\ \hline
\multicolumn{1}{|l|}{Metric} & Roll ($\phi$) & Pitch($\theta$) & Yaw ($\psi$) & \cellcolor[HTML]{DAE8FC}Average      & Roll ($\phi$) & Pitch($\theta$) & Yaw ($\psi$) & \cellcolor[HTML]{DAE8FC}Average      \\ \hline
\multicolumn{1}{|l|}{MAE}    & 2.89          & 1.52            & 4.07         & \cellcolor[HTML]{DAE8FC}2.83         & 3.90          & 5.26            & 3.42         & \cellcolor[HTML]{DAE8FC}4.20         \\
\multicolumn{1}{|l|}{MSE}    & 23.23         & 5.59            & 27.2         & \cellcolor[HTML]{DAE8FC}18.67        & 34.81         & 45.59           & 20.55        & \cellcolor[HTML]{DAE8FC}33.65        \\
\multicolumn{1}{|l|}{IAE}    & 2,887.55      & 1,522.62        & 4,071.63     & \cellcolor[HTML]{DAE8FC}2,827.26     & 3,904.83      & 5,258.45        & 3,423.00     & \cellcolor[HTML]{DAE8FC}4,195.42     \\
\multicolumn{1}{|l|}{ISE}    & 23,226.56     & 5,588.60        & 27,203.42    & \cellcolor[HTML]{DAE8FC}18,672.86    & 34,811.32     & 45,589.85       & 20,548.85    & \cellcolor[HTML]{DAE8FC}33,650.01    \\
\multicolumn{1}{|l|}{ITAE}   & 179,944.50    & 93,339.33       & 261,947.22   & \cellcolor[HTML]{DAE8FC}178,410.35   & 236,408.09    & 320,204.59      & 217,343.01   & \cellcolor[HTML]{DAE8FC}257,985.23   \\
\multicolumn{1}{|l|}{ITSE}   & 1,499,075.54  & 369,576.98      & 1,893,954.11 & \cellcolor[HTML]{DAE8FC}1,254,202.21 & 2,100,576.25  & 2,927,031.10    & 1,419,390.55 & \cellcolor[HTML]{DAE8FC}2,148,999.30 \\ \hline
\end{tabular}
\caption{\new{Performance metric comparison of the simulation evaluation for the NN controller and PID controller. Metric is reported for each individual axis, along with the average. Lower values are better.}}
\label{tab:flight_metrics}
\end{table*}

As we can see there is an increase in error transferring from
simulation from reality, however this was expected because the digital twin does not perfectly
model the real system.
\new{The MAE is not significant 
however there is a large increase in error for the integral
    measurements.
    A partial explanation for this is if we refer to  {Fig.}~\ref{fig:flight} (particularly the pitch axis) we can
    see the controller is consistently off by about 10
    degrees which will continually add error to these measurements. }
The increased error on the pitch axis appears to be due to the
differences in frame shape between the digital twin and real
quadcopter, which are both asymmetrical but in relation to a different
axis.
This discrepancy may have resulted in pitch control
lagging in the real world as more torque and power is required to pitch in our
real quadcopter.   

A more accurate digital twin model can boost accuracy. 
Furthermore, during this particular flight wind gusts exceeded 30mph, while in the
simulation world there are no external disturbances acting upon the aircraft.
In the future we plan to deploy an array of sensors to measure wind speed so we
can correlate wind gusts with excessive error. 
Nonetheless, as shown in the video, stable flight can be maintained
demonstrating the transferability of a \nn trained with our approach.

\textbf{PID vs \nn Control.} Next we performed an experiment to
compare the performance of the \nn controller used in \fc to a PID
controller in simulation using the \newgym environment.  
\new{Although other control algorithms may exist in literature that out perform
    PID, of the open source flight controllers avail be for
    benchmarking, every single one uses PID~\cite{ebeid2018survey}.
    A major contribution of this work is providing the research community an
    additional flight control algorithm for benchmarking.
}

The PID
controller was first tuned in the simulator using the classical
Ziegler-Nichols method~\cite{ziegler1942optimum} and then manually
adjusted to \new{reduce overshoot} to obtained the following gains for
each axis of rotation: $K_\phi = [0.032029, 0,0.000396 ]$, $K_\theta =
[0.032029, 0, 0.000396 ]$, $K_\psi = [0.032029, 0, 0]$, where
$K_\text{axis} = [K_p,K_i,K_d]$ for each proportional, integrative,
and derivative gains, respectively.
\new{It took  approximately a half hour to manually tune the 9 gains with the
    bottleneck being the time to execute the simulator in order to obtain the
    parameters to calculate Ziegler-Nichols. In comparison to training a \nn}
\new{via PPO, there is not a considerable overhead difference given this is an
    offline task. In fact the tuning rate by PPO is significantly
    faster by a factor of 75.}

The RC commands from the real test flight where then replayed back to the
simulator similar to the previous experiment, however this time using the tuned PID controller. A 
\new{zoomed in comparison of
the \nn} \new{ and PID controller tracking the desired angular velocity for
two aerobatic maneuvers is
shown in Fig.~\ref{fig:zoom}. Although the performance is quite close,
we can most visibly the \nn} \new{controller tracking  the pitch axis during a
Split-S maneuver more accurately. }

\new{We also computed the same control measurements for the PID controller and
reported them in Table~\ref{tab:flight_metrics}. Results show, on average, the
\nn} \new{controller to out perform the PID controller for every one of our metrics. }

It is important to note PID tuning is a challenging task and the PID
controller's accuracy and ability to control the quadcopter is only as
good as the tune. The \nn controller on the other hand did not require
any manually tuning, instead through RL and interacting with the
aircraft over time it is able to teach itself attitude control.  As
we continue to the reduce the gap between simulation and the real
world, the performance of the \nn controller will continue to improve
in the real world.

\section{Future Work and Conclusion}
\label{sec:con}
In this work we introduced \fc, the first open-source neuro-flight control
firmware for remote piloting multi-copters and fixed wing aircraft and its
accompanying toolchain. %
There are \new{four} main directions we plan to
pursue in future work.
\begin{enumerate}

\item \textbf{Digital twin development.} In this work we synthesized our \nn
using an existing quadcopter model that did not match NF1. Although stable flight was achieved demonstrating the \nns
robustness, comparison between the simulated flight verse the actual flight
is evidence inaccuracies in the digital twin has a negative affect in flight control accuracy.  
In future work we will develop an accurate digital twin of NF1
and investigate how the fidelity
of a digital twin affects flight performance in an effort to reduce costs during
development. 
\item \textbf{Adaptive and predictive control.} With a stable platform in place
we can now begin to harness the \nn's true potential.  We will enhance the
training environment to teach adaptive control to account for excessive sensor noise, voltage
sag, change in flight dynamics due to high throttle input, payload changes, external disturbances such as
wind, and propulsion system failure. 
\item \textbf{Continuous learning.} Our current approach trains \nns exclusively using offline learning. However, in order to 
reduce the performance gap between the simulated and {real world},
we expect that a hybrid architecture involving online incremental
learning will be necessary. % to provide continuouslearning.
Online learning will allow the aircraft to adapt, in real-time, and
to compensate for any modelling errors that existed during synthesis of
the \nn during offline (initial) training.
Given the payload restrictions of micro-UAVs and weight associated
with hardware necessary for online learning we will investigate
methods to off-load the computational burden of incremental learning
to the cloud.
\item \textbf{\nn architecture development}
    \new{ Several performance benefits can be realized from an optimal
         network architecture for flight control including improved
         accuracy~(Section~\ref{sec:flighteval}) and
         faster execution~(Section~\ref{sec:timing}). In future work we plan to
         explore recurrent architectures utilizing long short-term memory~(LSTM)
         to improve accuracy. Additionally we will investigate alternative
         distributions such as the beta function which is naturally bounded~\cite{chou2017improving}.
         Furthermore we will explore the use of the rectified linear unit~(ReLU) activation functions to increase execution
         time which is more computationally efficient than the hyperbolic tangent 
         function. }
\end{enumerate}
The economic costs associated with developing neuro-flight control
will foreshadow its future, determining whether its use will remain
confined to special purpose applications, or if it will be adopted in
mainstream flight control architectures.
Nonetheless, we strongly believe that \fc is a major milestone in
neuro-flight control and will provide the required foundations for
next generation flight control firmwares.

\balance
\bibliographystyle{IEEEtran}

\bibliography{references}

\end{document}